# FGSI: Distant Supervision for Relation Extraction method based on Fine-Grained Semantic Information

Chenghong Sun[1], Weidong Ji[1] [*], Guohui Zhou[1], Hui Guo[1], Zengxiang Yin[1] and Yuqi Yue[1]

1 College of Computer Science and Information Engineering, Harbin Normal University, Harbin 150025, China

**Corresponding author**: Weidong Ji (e-mail: kingjwd@126.com)
**Contributing authors**: Chenghong Sun (e-mail: schong@stu.hrbnu.edu);
Guohui Zhou(e-mail: zhouguohui@hrbnu.edu.cn);
Hui Guo(guohui_hnu@163.com);
Zengxiang Yin(yinzengxiang@163.com);
Yuqi Yue(yueyuqi980717@163.com)

**ABSTRACT**: Relation extraction is an important part of constructing a knowledge graph, whose main purpose is to extract the semantic relationships between marked entity pairs in a sentence. It plays a crucial role in understanding the semantics of a sentence and constructing a knowledge graph. Remote supervision relation extraction methods alleviate the pressure of manually constructing datasets by automatically aligning knowledge bases with natural language texts to generate labeled data. However, the labeled data obtained by remote supervision often contains a large amount of noisy data, which can significantly affect the training of relation extraction models. This paper proposes the hypothesis that key semantic information within a sentence plays a critical role in entity relation extraction tasks and addresses the issue of noise in remote supervision relation extraction tasks. Based on this hypothesis, the paper splits a sentence into three segments according to the position of entities and uses intra-sentence attention mechanism to identify fine-grained semantic features within the sentence and reduce the interference of irrelevant noise information. The paper then uses inter-package attention mechanism to increase the weight of correctly labeled sentences and discard noisy sentences, thereby minimizing the impact of noise on the relation extraction model and fully utilizing existing positive semantic information. Experimental results show that the proposed relation extraction model improves precision-recall curve and P@N values compared to existing methods, demonstrating the effectiveness of the model.

**Keywords**: Relation Extraction; Distant Supervision; Fine-grained Semantic Information; Attention Mechanism; Threshold Gate

# 1 Introduction

Relationship extraction aims to identify the relationship between entity pairs in plain text sentences to obtain structured knowledge information, i.e., triple information (Entity A, Relation, Entity B), which is an important research hotspot in natural language processing[1] and an essential preparatory work for constructing knowledge graphs [2]. Currently, machine learning methods for relationship extraction can be divided into unsupervised learning [3], supervised learning [4], semi-supervised learning [5], and remote supervision learning [6] according to whether the required training corpus is annotated. Although supervised learning methods for relationship extraction have high accuracy and satisfactory overall performance, they require manual annotation of the dataset before model training, which involves a significant amount of human, material, and financial resources. With the continuous development of relationship extraction technology, Mintz [6] et al. proposed the idea of remote supervision in 2009, which automatically aligns the knowledge base with plain text to generate annotated data. The main idea is based on a strong assumption that "if two entities have a certain relationship in the knowledge base, then all sentences containing these two entities will express this relationship." For example, (Huawei, founder, Ren Zhengfei) is a triple relationship instance in Freebase, and all sentences containing these two entities will be labeled as founder relationship. However, the remote supervision method proposed by Mintz [6] et al. still has flaws. The strong assumption they proposed for relationship extraction tasks leads to incorrect annotation problems in the generated dataset, resulting in noise interference in the actual model training process and affecting model performance.

One of the main research directions for distant supervision relation extraction is to develop denoising methods for the relation model, as proposed by Yang Suizhu et al. [7]. In recent years, scholars have proposed various solutions for sample denoising. Surdeanu et al. [8] addressed the noisy label problem by adopting a multi-instance learning strategy. Takamatsu et al. [9] designed a generative model to identify patterns of positive and negative samples, discarding negative pattern samples and retaining positive pattern samples to improve the overall performance of the relation extraction model. Zeng et al. [10] considered the limitations of traditional natural language processing tools and proposed the use of convolutional neural networks for relation extraction, using word vectors and word position vectors as inputs, which achieved better results than classical machine learning models. Nguyen et al. [11] proposed using windows of multiple scales to extract multidimensional features instead of conventional lexical features, which achieved better results than traditional convolutional neural network models. Zeng et al. [12] designed a segmented convolutional neural network to extract sentence features and used multi-instance learning to eliminate annotation errors in incorrect samples, reducing the impact of erroneous samples on the overall model performance. Yan Xu et al. [13] first proposed using Long Short-Term Memory (LSTM) networks for relation extraction and extracted key information through the shortest dependency path, enabling better extraction of sentence-level relations. Yankai Lin et al. [14] improved the selection of training sentences in each bag of multi-instance learning by designing a bag-level attention mechanism to score all sentences in the bag and integrate all sentence information for relation extraction, achieving better results than the baseline model. Guoliang Ji et al. [15] introduced entity description information and sentence-level attention mechanism for distant supervision relation extraction, further enriching entity information, reducing noise interference, and achieving better results than previous baseline models. Peng Zhou et al. [16] proposed using hierarchical selective attention for distant supervision relation extraction, where coarse sentence-level attention was used to select relevant sentences, word-level attention was used to construct sentence representations, and fine-grained sentence-level attention was used to aggregate sentence representations as model inputs, demonstrating the superior performance of their model through experiments. Feng Jianzhou et al. [17] proposed an improved attention mechanism for relation extraction, in which the model found all positive instances that reflected the relation between the same entity pair at the sentence level, then constructed a combined sentence vector to fully utilize the semantic information of positive instances, achieving higher accuracy than the compared model. Ye Yuxin et al. [18] hypothesized that "the label of the final sentence alignment is a noisy observation result generated based on some unknown factors." They learned the transition probability from noisy labels to true labels by training on automatically labeled data for relation extraction, achieving better results than mainstream baseline models. Although sentence-level or bag-level attention mechanisms can achieve the goal of obtaining positive corpus, they do not consider the fine-grained semantic information within the sentence. If there is too much noise interference within the positive corpus, the corpus may be considered false positive by the program due to its low weight after attention calculation. This is catastrophic for distant supervision datasets with a large number of noisy sentences.

To accurately identify the relationship between two entities in a sentence, we need to focus on the semantic information within the sentence. A complete sentence typically consists of components such as subject, predicate, object, and adverbial. If a sentence can semantically express the relationship between two entities, it must be related to the key semantic information in the sentence, while other information is considered irrelevant or interfering noise. Liu et al.'s study [19] showed that in the classic dataset of distant supervised relation extraction, NYT-Freebase, nearly 99.4% of sentences contain a large amount of noisy words. If the entire sentence is input into the model for training without processing the fine-grained semantic features, it will inevitably be affected by irrelevant noise within the sentence, thus affecting the overall performance of the model.

This paper proposes a remote supervision relationship extraction model based on fine-grained semantic information piecewise convolutional neural networks (PCNN+FGSI). The main contributions of this paper are as follows: (1) a new intra-sentence attention mechanism is proposed, which is different from the coarse-grained attention mechanism established at the sentence level. It is used to process fine-grained semantic features within the sentence, highlighting key semantic information and preventing irrelevant information and noise information from participating in the construction of sentence feature vectors with the same weight; (2) Based on (1), after obtaining sentence features that highlight fine-grained semantic information, a bag-level attention mechanism is used to screen positive training sentences with threshold gates and discard noisy sentences, in order to better distinguish positive and negative instances within all sentences containing the same entity pair and construct a combination feature vector to train the relationship classification network; (3) Comparative experiments and ablation experiments are designed to verify the performance advantages of the proposed relationship extraction method.

# 2 Segmented convolutional neural network models based on fine-grained semantic information

This paper proposes a fine-grained semantic information piecewise convolutional neural network model (**PCNN+FGSI**) for remote supervised relation extraction. The entire model consists of four parts, namely the text embedding layer based on fine-grained semantic information, the single-sentence feature output layer, the multi-sentence combined feature output layer, and the relation classification layer. The overall structure of the model is shown in Figure 1.

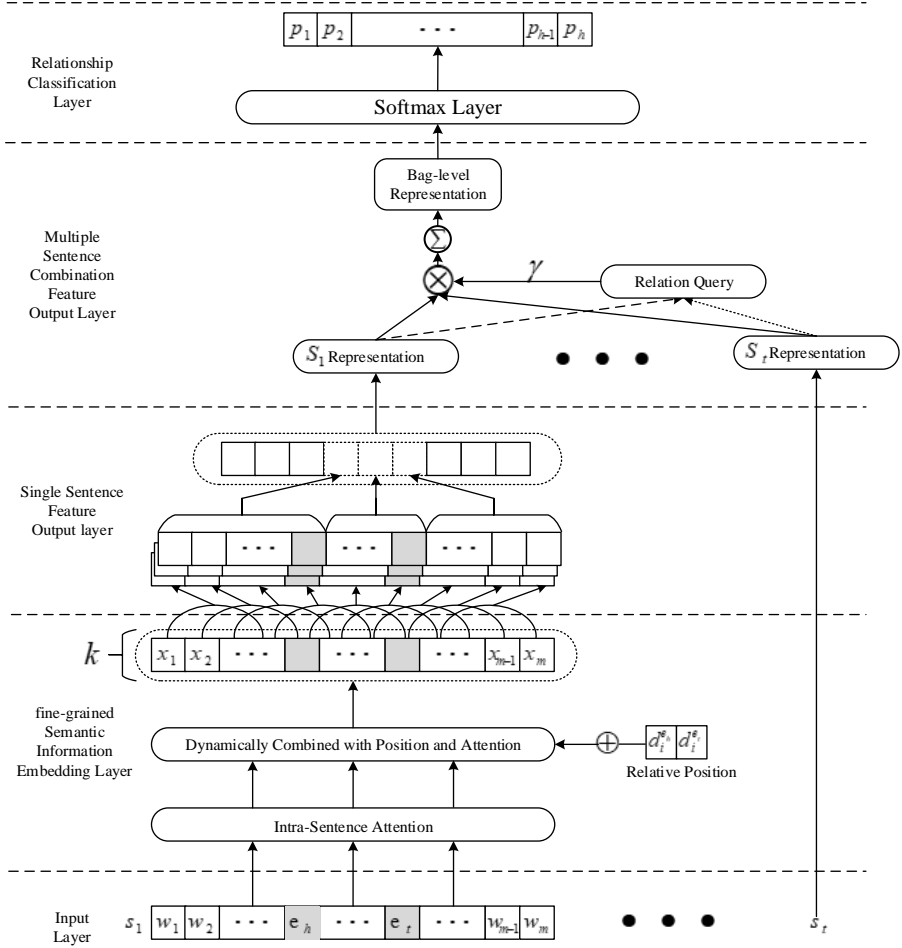

Figure 1. Overall architecture of the PCNN+FGSI model.

In the text embedding layer based on fine-grained semantic information, the entire sentence is divided into three parts based on the positions of the two entities, and then the intra-sentence attention mechanism is applied to increase the weight of the part containing key semantic information and decrease the weight of the part containing noise information. The resulting representation emphasizes fine-grained semantic information. After obtaining the semantic embedding representation that emphasizes fine-grained semantic information, the single-sentence feature representation is formed through the encoding layer. The package-level attention mechanism in the multi-sentence combined feature output layer is used to screen positive instance information from the sentence feature representations containing the same entity pair. The weights of the positive instance feature representations are obtained and then the feature vectors are recombined. The recombined feature vectors are sent to the relationship classification layer to train the classifier, which improves the training performance of the model.

## 2.1 Text embedding layer based on fine-grained semantic information

The proposed model relies on neural networks to accomplish the task of relation extraction. However, natural language text cannot be directly used by neural networks. Therefore, when completing natural language processing tasks with neural networks, the first step is to convert the natural language text into a real-valued vector representation. The based on fine-grained semantic information text embedding layer of this model processes natural language text in three steps, namely word embedding, intra-sentence attention mechanism, and relative position information embedding. The structure of the based on fine-grained semantic information text embedding layer is shown in Figure 2. After the training corpus is embedded by the word embedding part, the key semantic information part is given a greater weight by the intra-sentence attention mechanism, and then the relative position embedding information is concatenated to form the embedding vector representation of the sentence.

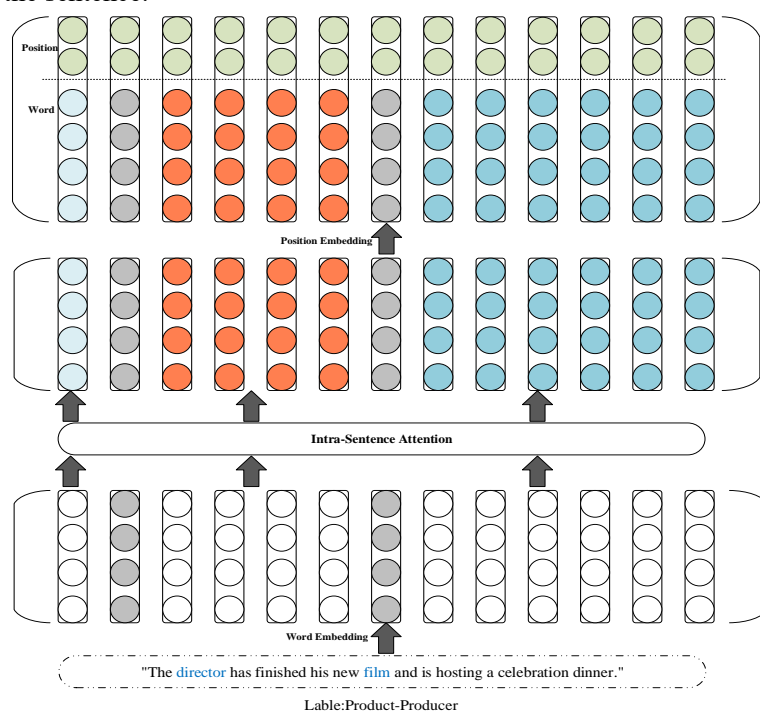

Figure 2: Structure diagram of the text embedding layer based on fine-grained semantic information.

### 2.1.1 Word Embedding

Word embedding is the process of transforming words into computable vectors, which are low-dimensional distributed representations of each word. The effectiveness of word embeddings in many natural language processing tasks has been demonstrated by Socher et al. [20]. Different methods have been proposed to train word embeddings, such as those by Bengio et al. [21] and Mikolov et al. [22]. Currently, the most commonly used pre-trained word vectors are LSA (Latent Semantic Analysis), Word2vec, and GloVe. LSA is an early count-based word vector representation tool based on co-occurrence matrix. It uses matrix factorization techniques based on singular value decomposition (SVD) to reduce dimensionality of large matrices. However, the computational cost of SVD is high. Word2vec's major limitation is that it only utilizes the corpus within a fixed window and does not fully leverage all the available corpus. GloVe combines the advantages of both methods. Figure 3 shows the distribution of the top 100 words with cosine similarity to the word "founder" in the semantic space of GloVe.

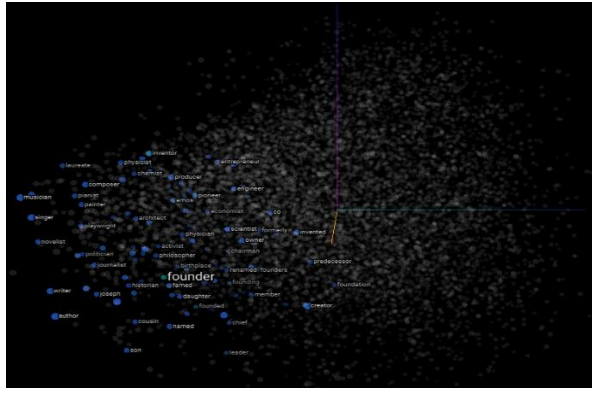

Figure 3: Distribution of Semantics in Space

In this model, we use the pre-trained word embeddings method from Stanford GloVe. Given a sentence $s = (w_1, w_2, w_3, e_h, w_5, \cdots, w_l, e_t, w_{l+2}, \cdots, w_m)$, each word is represented as a $k_w$-dimensional real-valued vector using the pre-trained word embedding matrix $\boldsymbol{E} \in \boldsymbol{R}^{|V| \times k_w}$, where $e_h$ and $e_t$ represent the head and tail entities, respectively. V is the size of the vocabulary (the number of words in the pre-trained word embedding corpus), and $m$ is the length of the sentence.

### 2.1.2 Intra-sentence attention mechanism

Assuming a sentence $s = (w_1, w_2, w_3, e_h, w_4, \cdots, w_l, e_t, w_{l+2}, \cdots, w_m)$ contains an entity pair $< e_h, e_t >$ and is labeled with relation $r$, the word embedding vector representation $s'$ of the sentence can be obtained using Section 2.1.1, which is a matrix $\boldsymbol{W}^{m \times k_w}$, where $m$ is the number of words in the sentence and $k_w$ is the dimension of the word embedding. In this paper, the word embedding vector representation of the sentence $s' = \{s'_1, s'_2, s'_3\}$ is divided into three segments based on the positions of the two entities $< e_h, e_t >$ in the sentence. If a sentence can express the semantic relationship between its two internal entities, it must be related to key semantic information. After dividing the sentence into three parts according to the positions of the entities, the contributions of different parts to the model's ability to extract the correct entity relation are different. To enable the model to better understand the key semantic information that expresses different entity relations, different weights are assigned to these three parts to reflect their contribution to relation r. The equation for calculating the weight of each part is as follows:

$$\alpha_i = \frac{\exp(e_i)}{\sum_{k=1}^{3} \exp(e_k)}, 1 \leq i \leq 3 \tag{1}$$

Where $e_i$ is the contribution of the i-th segment of the sentence to the relation label $r$ after the sentence is divided into three parts, and the calculation equation is as follows:

$$e_i = \frac{s'_i \cdot r'}{\|s'_i\| \times \|r'\|} = \frac{\sum_{j=1}^{k_w} s'_{ij} \times r'_j}{\sqrt{\sum_{j=1}^{k_w} (s'_{ij})^2} \times \sqrt{\sum_{j=1}^{k_w} (r'_j)^2}} \tag{2}$$

Where $s'_i$ represents the embedded vector representation of the $i$-th part of the sentence after embedding, and $r'$ represents the embedded vector representation of the relationship label $r$ in the semantic space used by this model. After calculating the contribution of each part, the formula for calculating the final embedded vector of the sentence is as follows:

$$s' = [a_1 s'_1; a_2 s'_2; a_3 s'_3] \tag{3}$$

### 2.1.3 Position Embedding

Zeng et al. [10] have shown through experiments the importance of positional features in relation extraction tasks. Feng et al. [17] also argue that when judging the relationship between entity pairs in a sentence, words that are closer to the entities are usually key information. Therefore, in order to better capture the structural information of a sentence, this paper introduces positional embeddings in the embedding stage, using positional features to record the relative distances of each word to the two entities. An example of relative distances is shown in Figure 4.

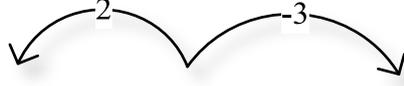

The [director] has finished his new [film] and is hosting a celebration dinner.

Figure 4: Example of Relative Distance

The model looks up the relative distance of each word $w_i$ to the two entities, and then maps these two relative distances to two $k_p$-dimensional real-valued vectors $(d_i^{eh}, d_i^{et})$. For each sentence that needs to be trained in the model, its word embedding and position embedding are concatenated to obtain the sentence vector representation matrix $X = [x_1, x_2, \cdots, x_m] \in R^{m \times k}$, where $x_i = [w_i; d_i^{eh}; d_i^{et}]$, $m$ denotes the length of the sentence, $k$ is the dimension after concatenating the word embedding and position embedding vectors, that is, $k = k_w + k_p \times 2$.

## 2.2 Single-sentence feature output layer

The effectiveness of the PCNN model for sentence-level feature extraction has been demonstrated in the studies by Zeng et al. [10] and G. Ji et al. [15]. Therefore, in this paper, we adopt the PCNN structure as the single-sentence feature output layer of our model, as shown in Figure 5. After obtaining the embedded representation of the sentence, the embedding vector is fed into the PCNN structure, and the sentence's feature vector representation is obtained through convolutional and piecewise max-pooling computations.

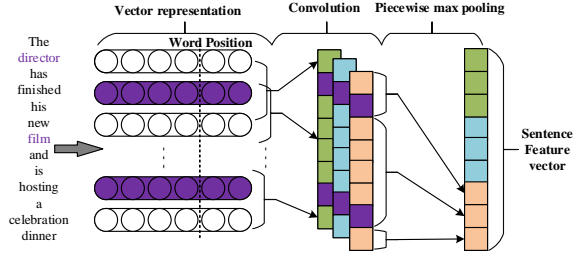

Figure 5: Network structure of PCNN

### 2.2.1 Convolution

In the task of entity relation extraction, the length of each sentence varies. To address this issue, sentence padding is applied to align the length of the corpus. The alignment standard is based on the longest sentence in each $batch$ of samples. Additionally, effective information for determining the relationship between target entities may exist at different positions within a sentence. To capture such information from different positions, the model needs to extract local features at different scales to predict the relationship classification for the entity pair. In deep learning, the convolution operation is often used to extract local features of different scales. Dumoulin V et al. [24] conducted in-depth research on convolution algorithms in deep learning.

After the text is embedded with the fine-grained semantic information in the text embedding layer,

the final embedding vector representation of the input sentence is defined as $s'' = \{b_1, b_2, \cdots, b_{|s''|}\}$, where $b_i$ denotes the embedding vector representation of the $i$-th word in the sentence, and $b_i \in R^{|k|}$. In this paper, $s''_{i:j}$ is used to represent the horizontal concatenation matrix of the embedding sequence $[b_i, b_{i+1}, \cdots, b_j]$ in the sentence, and $w$ represents the length of the filter operator. The weight matrix of the filter operator is denoted as $W \epsilon R^{w \times k}$. The convolution operation is performed by filtering the embedding vector representation of the sentence with the filter operator, and a vector $c \in R^{|s''|-w+1}$ is obtained, as shown in equation (4):

$$c_j = W \otimes s''_{(j-w+1):j} \qquad (4)$$

In this formula, $1 \leq j \leq |s''| - w + 1$. During the feature extraction process through convolution, different filter kernels are needed to extract feature information at various positions in the sentence instance. Therefore, $n$ different filter kernels are used, and correspondingly, there are $n$ weight matrices $\widehat{W} = \{W_1, W_2, \cdots, W_n\}$. All convolution operations during the feature extraction process can be represented by equation (5):

$$c_{ij} = W_i \otimes s''_{(j-w+1):j} \qquad (5)$$

Here, $1 \leq i \leq n$ and $1 \leq j \leq |s''| - w + 1$. The convolution operation produces feature vectors for each sentence, denoted as $C = \{c_1, c_2, \cdots, c_n\}$.

### 2.2.2 Piecewise Max Pooling

The max-pooling operation selects only the strongest response in each feature map to pass to the next layer, while discarding other elements, which eliminates a lot of redundant information in the network and makes the entire network structure easier to optimize. However, the single max-pooling operation also has drawbacks, as it often loses some detailed information in the feature maps. In order to solve this problem and make the model more robust to important feature location changes, PCNN divides the sentence instances into three parts based on the positions of the two entities in the given sentence, and performs max-pooling on each part separately.

After the convolution operation in 2.2.1, the feature vector $c_i$ can be obtained, which can be represented as $c_i = \{c_{i\_1}, c_{i\_2}, c_{i\_3}\}$ by dividing the sentence instance into three parts according to the positions of the given entities. Based on this vector, the segmented max pooling operation is performed, i.e., $p_{ij} = \max(c_{i\_j})$, where $1 \leq i \leq n, j = 1,2,3$. Then, the resulting vectors are concatenated to obtain $p_i = [p_{i1}, p_{i2}, p_{i3}](i = 1,2,\cdots,n)$, where $p \epsilon R^{3n}$. This represents the feature vector of each sentence obtained after being processed by the PCNN structure.

### 2.3 Multilingual Sentence Combination Feature Output Layer

In order to automatically filter out noisy sentences with significant differences from the labels during the remote supervised relation extraction task, this layer adopts a multiple-instance learning strategy and an intra-bag attention mechanism. It discards all noisy sentences and combines the features of all positive instance sentences to form the training positive examples for the final classifier. The structure of this layer is shown in Figure 6. Each sentence feature vector within a bag is subjected to attention calculation with the relation query vector, and the corresponding weights are obtained. Sentences with weights lower than the hyperparameter β are filtered out by a threshold gate, and then the bag-level vector representation is formed and involved in the training based on the weights.

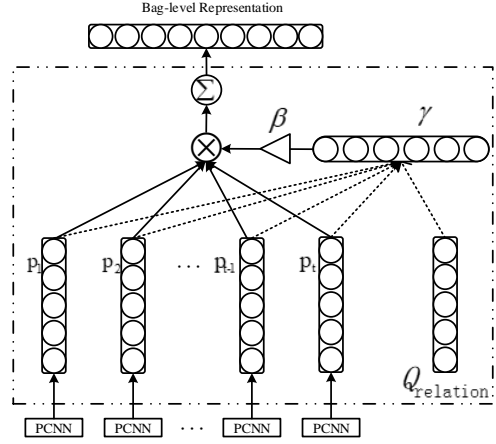

Figure 6: Multilingual Sentence Combination Feature Output Layer

This paper includes all sentences that contain the given entity pair $\langle e_1, e_2 \rangle$ and have a relationship label $r$ into the set $S$. Assuming that there are $t$ sentences that meet the requirement, the set $S$ can be represented as $S = \{s_1, s_2, s_3, \cdots, s_t\}$. After obtaining the feature vector representation $p$ for each sentence in Section 2.2, the vector set P corresponding to the sentence set $S$ can be represented as $P = \{p_1, p_2, p_3, \cdots, p_t\}$. Due to the noise problem in remote supervision, each sentence in this set expresses the relationship label $r$ differently. Therefore, an intra-bag attention mechanism is adopted to set a weight that can express the relationship label $r$ for each sentence through attention calculation. After threshold gating, the weights $(\gamma_1, \gamma_2, \gamma_3, \cdots, \gamma_n)$ for each sentence involved in the formation of the bag-level vector representation are calculated using the formula in equation (6):

$$\gamma_i = \frac{\exp(e_i)}{\sum_{k=1}^{n} \exp(e_k)}, 1 \leq i \leq n; e_i \geq \beta \tag{6}$$

Here, $e_i$ represents the relevance degree of the i-th sentence in the set $S$ to the relationship label $r$, and its calculation formula is shown in equation (7):

$$e_i = \frac{p_i \cdot Q_{relation}}{\|p_i\| \times \|Q_{relation}\|}, 1 \leq i \leq t \tag{7}$$

Here, $p_i$ represents the feature vector of the i-th sentence in the sentence set $S$, and $Q_{relation}$ is the vector representation of the relationship label $r$ in the semantic space, representing the weight of the relationship label $r$ in calculating each sentence.

After the calculation of intra-sentence attention, each sentence in the set $S$ has obtained a weight that expresses the relationship label $r$. This paper believes that different sentences in the same set have different degrees of expression for the relationship label $r$, which can be reflected in the weight $\gamma$ accordingly. Therefore, positive instances score high on weight $\gamma$, while negative instances score low on weight $\gamma$. Based on the above assumptions, by setting the hyperparameter $\beta$, when forming the combination feature vector of multiple sentences, the sentence vectors with weights lower than $\beta$ are filtered out, thus avoiding noise sentences from participating in the formation of combination feature vectors with low weights. Assuming that after filtering out noise sentences, there are still $n$ sentences left in the set $S$, the formula for generating the combination feature vector of the set is shown in equation (8):

$$g = \sum_{j=1}^{n} \gamma_i p_j, 1 \leq j \leq n \tag{8}$$

## 2.4 Relation Classification Layer

For the set $S$ in Section 2.3, where the distant supervision relationship label is known, in order to compute the probability distribution of the combined feature vector of the set for relationship classification, the $softmax$ layer is applied to the relationship classification layer in this paper. Assuming that the combined feature vector of the $i$-th set $S$ is denoted as $g_i$, the probability distribution of the relationship obtained by passing the combined feature vector through the softmax layer is shown in equation (9):

$$P(r_i|g_i) = softmax(W_o g_i + b_o) \tag{9}$$

Here, $W_o \epsilon R^{h \times 3n}$, where $h$ represents the number of pre-defined relations.

## 2.5 Optimization

The model parameters to be optimized in this paper are $\theta = (E, D_{he1}, D_{te2}, W, W_o)$, where $E$ represents the word embeddings, $D_{he1}$ represents the position vectors of words relative to the head entity, $D_{te2}$ represents the position vectors of words relative to the tail entity, $W$ represents the parameters involved in the convolutional operation, and $W_o$ represents the parameters of the relation classification layer. The cross-entropy loss function used in this model is defined as shown in equation (10):

$$J(\theta) = \sum_{i=1}^{N} \log p(r_i|g_i, \theta) \tag{10}$$

Where $N$ is the number of sentence sets, and $g_i$ represents the combined feature vector of the $i$-th sentence set.

During parameter updates, Li et al. [25] compared four common optimizers by performing parameter optimization on the hand-written digit recognition MNIST dataset and the FASHION dataset. Among them, the $Adam$ optimizer performed well. Therefore, the $Adam$ optimizer was used as the parameter update optimizer for the model in this paper. The Adam optimizer combines the first-order moment of the gradient of SGD-M and the second-order moment of the gradient of RMSprop, taking into account the mean and variance of the gradient, and adds two correction terms on this basis. The formula is shown in equations (11)-(13):

$$m_t^1 = \frac{m_t}{1-\beta_1^\tau} \tag{11}$$

$$v_t^2 = \frac{v_t}{1-\beta_2^\tau} \tag{12}$$

$$\omega_{t+1} = \omega_t - lr \times \frac{m_t^1}{v_t^2} \tag{13}$$

Here, $m_t^1$ represents the bias-corrected first moment estimate and $v_t^2$ represents the bias-corrected second moment estimate, where $\beta_1, \beta_2 \in [0,1]$ are the decay rates of the first and second moment estimates respectively, and $lr$ denotes the learning rate.

# 3 Experimentation and Evaluation

To demonstrate the effectiveness of the proposed method in this paper, comparative experiments and ablation experiments were designed in this section to demonstrate the advantages of the proposed method from different perspectives.

## 3.1 Dataset and Evaluation Metrics

The NYT-10 dataset was released by Riedel et al. [12], and many domestic and foreign scholars have conducted research on distant supervision relation extraction based on this dataset [27][32]. The dataset is aligned with relations in Freebase, and the sentences obtained from news corpus from 2005 to 2006 are used as the training set, while the sentences obtained from news corpus in 2007 are used as the test set. The dataset contains 53 types of relations, including the special relation type "NA", which indicates that there is no relation between two entities. In both the training and test sets, the special relation type "NA" has the largest proportion among all the training sentences. We set the maximum length of sentences in the dataset to 256, and Figure 7 shows the distribution of sentence lengths in the NYT-10 dataset. It can be seen that the maximum length of sentences is concentrated within [20, 60].

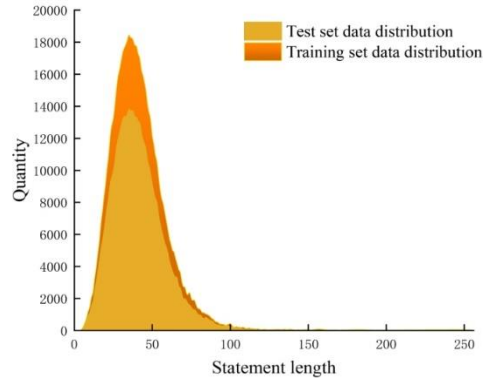

Figure 7: Data distribution of NYT-10 dataset

We use the held-out evaluation method to evaluate the proposed relation extraction model, and evaluate the performance of the model through the $PR(Precision-Recall)$ curve and $P@N(Precision@Top\ N)$.

## 3.2 Parameter Settings

In this study, we tested the performance of the model on the test dataset by adjusting parameters such as the maximum length of training sentences, polynomial decay learning rate, hyperparameters, and batch size. The other parameters were the same as those used by Lin et al. [26]. Table 1 shows the main parameters used in the experiments of this study.

Table 1 Parameter Settings

| Parameter Description | Configuration |
|---|---|
| Convolutional Kernel Size | 3、4、5 |
| Number of Convolutional Kernels | 200 |
| Word Embedding Dimension | 200 |
| Positional Embedding Dimension | 5 |
| Batch Size | 128 |
| Dropout | 0.5 |
| Initial Learning Rate Minimum Learning Rate | 1E-2 |
| Initial Learning Rate Minimum Learning Rate | 1E-6 |

### 3.3 Comparative experimental results and analysis

To evaluate the proposed method on the NYT-10 dataset, we selected several classic baseline methods for comparison through held-out evaluation. The compared baseline methods are:

- Mintz [6]: Mintz first proposed the idea of distant supervision and combined the advantages of supervised and unsupervised information extraction.
- MultiR [28]: This model, proposed by Hoffmann et al., combines a sentence-level extraction model with a simple corpus-level component for aggregating single facts.
- MIML [8]: This is a multi-instance multi-label learning method proposed by Surdeanu.
- PCNN+MAX [10]: This method, proposed by Zeng, trains instances with the maximum logistic regression value.
- PCNN+ATT (Sentence-level Selective Attention Model) [26]: This is an improved model based on the PCNN model, proposed by Lin et al., which uses sentence-level attention mechanism.
- PCNN+MIL [10]: This method, proposed by Zeng, combines the advantages of multi-instance learning and the PCNN model.
- PCNN+RL [31]: This method, proposed by Jun Feng et al., applies reinforcement learning to instance selectors to choose high-quality sentences for training the relation classifier.
- APCNNS [15]: This is an extraction method that combines PCNN with entity information, proposed by Ji.
- BGWA [29]: This method, proposed by Jat S et al., uses word-level attention mechanism for relation extraction tasks.

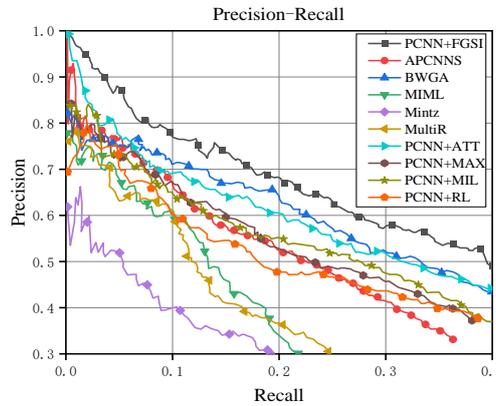

- Figure 8: Precision-Recall Curve

From Figure 8, it can be observed that: (1) Traditional relation extraction methods, such as Mintz and MultiR, are feature-based research methods that do not consider the issue of noise in the dataset. Although they can complete relation extraction tasks by extracting features, the experimental results are relatively poor due to the interference of too much noise, which further highlights the importance of extracting features through denoising from large-scale data. (2) In the models using sentence-level attention, the experimental results of PCNN+RL, PCNN+MAX, and PCNN+MIL are inferior to that of the PCNN+ATT model because the PCNN+ATT model not only makes full use of the information provided by multiple instance sentences but also lowers the weight of noise sentences to reduce their negative impact during the entire model training period. (3) The APCNNS model for relation extraction, which provides additional entity information, enriches instance features and compensates for noise interference using external knowledge, resulting in an improvement compared to the PCNN+RL model. (4) The proposed PCNN+FGSI model avoids the participation of noise sentences in the feature vector through threshold gating, thereby avoiding interference from noise sentences. Compared with the BGWA model, the proposed model not only focuses on a single word but also considers the segment composed of multiple words. By judging the semantics of the segment and then magnifying the important semantic information of the segment through intra-sentence attention, the internal noise of the sentence instances involved in the training is further reduced. Across the entire recall range, the PCNN+FGSI model proposed in this paper achieved the highest precision.

Table 2 shows the comparison of P@N values between the proposed relation extraction method and baseline models. As can be seen from the table, among all the baseline models, the BGWA model has the slowest precision decline. Although the proposed PCNN+FGSI model does not perform as well as the BGWA model in terms of the rate of precision decline, it performs the best within the scope of the indicators. The average precision of PCNN+FGSI model is 8 percentage points higher than that of the PCNN+ATT model, which further validates the advantages of the proposed method.

Table 2 P@N comparison table of PCNN+FGSI and baseline model

| Methods | P@N(%) | | | |
|---|---|---|---|---|
| | 100 | 200 | 300 | Average |
| Mintz | 54.0 | 50.5 | 45.3 | 49.9 |
| MIML | 70.9 | 62.8 | 60.9 | 64.9 |
| MultiR | 64.0 | 61.5 | 53.7 | 59.7 |
| PCNN+MAX | 73.3 | 70.3 | 65.3 | 69.6 |
| PCNN+ATT | 81.1 | 71.1 | 69.4 | 73.9 |
| PCNN+MIL | 74.3 | 71.7 | 66.1 | 70.7 |
| PCNN+RL | 74.8 | 68.2 | 61.9 | 68.3 |
| APCNNS | 76.3 | 74.2 | 69.4 | 73.3 |
| BGWA | 75.2 | 74.1 | 71.4 | 73.6 |
| PCNN+FGSI | **86.5** | **82.7** | **76.4** | **81.9** |

## 3.4 Ablation experiment results and analysis

To investigate the role of the fine-grained semantic information in the text embedding layer in the model experiments, this section designs an ablation experiment. The control group (CG) represents the proposed model (PCNN+FGSI), while the experimental group (EG) uses a regular text embedding layer. Figure 9 shows the PR curves of the experimental group and the control group.

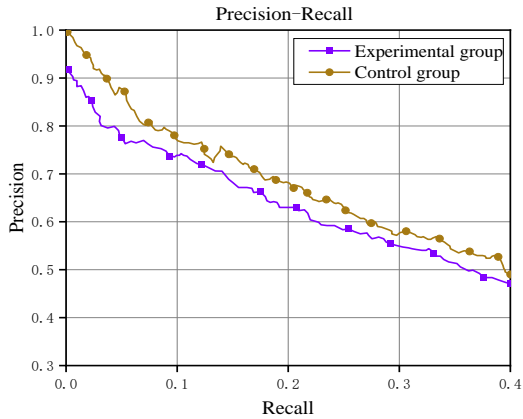

Figure 9: PR curve of the control experimental group

From Figure 9, it can be seen that the control group performs the best on the PR curve. The performance of the experimental group drops slightly when using a regular text embedding layer. This is because the text embedding layer based on fine-grained semantic information can highlight the semantic information that expresses entity relationships in positive instances, enabling the model to learn fine-grained semantic information that expresses entity relationships, and thereby constructing more robust feature vectors.

This paper also uses P@N to compare the performance of the experimental group and the control group, as shown in Table 3.

Table 3 P@N comparison table of experimental group and control group

| Methods | P@N (%) | | | |
| --- | --- | --- | --- | --- |
| | 100 | 200 | 300 | Average |
| EG | 82.3 | 78.4 | 74.2 | 78.3 |
| CG | 86.5 | 82.7 | 76.4 | 81.9 |

From Table 3, it can be observed that the experimental group with regular text embedding layer shows a decrease in performance in the P@N (N=100/200/300) evaluation metrics compared to the control group. This is consistent with the conclusion obtained from the PR curve analysis, indicating that the text embedding layer based on fine-grained semantic information is helpful in improving the model performance.

## 4 Conclusion

This paper proposes a relation extraction model based on fine-grained semantic information for distant supervision. In order to fully reduce the interference of noisy data in the distant supervision dataset and make full use of fine-grained semantic information in training instance sentences, the model first starts from within the sentence to find fine-grained semantic information that can reflect the label relationship in the sentence segment, reducing the interference of irrelevant semantic information, and forming a single-sentence feature vector. Then, by calculating the relevance of each sentence to the label within the same bag and screening positive instances through a threshold gate, all noise sentences are discarded, forming a high-quality combination feature vector to train the classifier. A large number of experiments have shown that the proposed model outperforms the baseline models on P@N and other indicators.

## ADDITIONAL INFORMATION

**Declaration of competing interest**

The authors declare that we have no known competing financial interests or personal relationships that could have appeared to influence the work reported in this paper.

**Acknowledgment**

We would like to express our heartfelt thanks to all those who have contributed to the completion of this research project.

First and foremost, we are deeply grateful to our supervisor for his guidance and support throughout the entire research process. We appreciate his invaluable advice, encouragement, and motivation, which have been instrumental in shaping our research direction and methodology.

We would also like to extend our sincere gratitude to our colleagues who have offered their assistance and collaboration, as well as to the participants who have generously given their time and effort to make this study possible.

Last but not least, we would like to acknowledge the funding agencies that have supported this research financially. Their contributions have enabled us to carry out this study and achieve our research goals.

Once again, we express our heartfelt thanks to all those who have contributed to this project, and we hope that our research will contribute to the advancement of knowledge in our field.

**Funding**

This research was funded by the National Natural Science Foundation of China, grant number 31971015 and funded by Natural Science Foundation of Heilongjiang Province in 2021 under, grant no LH2021F037.